\documentclass[conference]{IEEEtran}
\IEEEoverridecommandlockouts
\usepackage{cite}
\usepackage{amsmath,amssymb,amsfonts}
\usepackage{algorithmic}
\usepackage{graphicx}
\usepackage{textcomp}

\usepackage{graphicx}
\usepackage{graphics} % for pdf, bitmapped graphics files
\usepackage{epsfig} % for postscript graphics files
\usepackage{subfigure}
\usepackage{epstopdf}

\usepackage{pict2e}
\usepackage{tikz}
\usepackage{amsmath}

\usetikzlibrary{shapes,arrows}
\usepackage{amsmath,bm,times}
\usepackage{graphicx,adjustbox}
\usetikzlibrary{shapes,arrows}
\usetikzlibrary{positioning}
\usetikzlibrary{arrows,decorations.markings}

\usepackage{graphics} % for pdf, bitmapped graphics files
\usepackage{epsfig} % for postscript graphics files
\usepackage{subfigure}

\usepackage{lipsum}% http://ctan.org/pkg/lipsum %% WHY :
\usepackage{lipsum,adjustbox}
\usepackage{arydshln}
\usepackage{adjustbox}% To adjust tike picture size
\usepackage[shortcuts,acronym]{glossaries}
\makeglossaries % generates the acronym list
\usepackage{multicol}

\usepackage{amssymb,amsmath}
\usepackage{relsize}
\usepackage{tikz-3dplot}
\usepackage[english]{babel}
\usepackage{filecontents}
\usepackage{bm}
\usepackage{subfigure}
\usepackage{gensymb}
\usepackage{graphicx,adjustbox}

\usepackage{float}
\graphicspath{{Fig/}{Figures/}}

\usepackage{calrsfs}
\DeclareMathAlphabet{\pazocal}{OMS}{zplm}{m}{n}

\usepackage{algorithm}
\usepackage{algorithmic}
\usepackage{amsmath} % assumes amsmath package installed
\usepackage{amssymb}  % assumes amsmath package installed
\usepackage{graphicx} % for pdf, bitmapped graphics files
\usepackage{amsfonts}
\usepackage{xcolor}
\usepackage{cite}
\usepackage{subfigure}
\usepackage{amsfonts}
\usepackage{xr}
\usepackage{hyperref}
\usepackage{mathtools}

\renewcommand{\vec}[1]{{\boldsymbol{#1}}}

\newcommand{\x}{\vec{x}}
\renewcommand{\u}{\vec{u}}

\renewcommand{\v}{\vec{v}}

\newcommand{\J}{\vec{J}}

\newcommand{\q}{\vec{q}}

\newcommand{\bomega}{\vec{\omega}}

\newcommand{\btheta}{\vec{\theta}}

\usepackage{comment}

\usepackage[printwatermark]{xwatermark}
\usepackage{xcolor}
\usepackage{graphicx}
\usepackage{lipsum}

\def\BibTeX{{\rm B\kern-.05em{\sc i\kern-.025em b}\kern-.08em
    T\kern-.1667em\lower.7ex\hbox{E}\kern-.125emX}}
\begin{document}

\title{Grasp that optimises objectives along post-grasp trajectories\\
{\small To be appeared in the Proceeding of ICRoM 2017}
\\
{}
\thanks{This project was funded by EU H2020 RoMaNS, 645582, and EPSRC EP/M026477/1. Stolkin was supported by a Royal Society Industry Fellowship.}
}

\author{\IEEEauthorblockN{1\textsuperscript{st} Amir M. Ghalamzan E.}
\IEEEauthorblockA{\textit{School of Metallurgy and Materials} \\
\textit{ University of Birmingham}\\
Birmingham, United Kingdom \\
a.ghalamzanesfahani@bham.ac.uk}
\and
\IEEEauthorblockN{2\textsuperscript{nd} Nikos Mavrakis}
\IEEEauthorblockA{\textit{School of  Metallurgy and Materials} \\
\textit{ University of Birmingham}\\
Birmingham, United Kingdom \\
nxm504@bham.ac.uk
}

\and
\IEEEauthorblockN{3\textsuperscript{rd} Rustam Stolkin}
\IEEEauthorblockA{\textit{School of Metallurgy and Materials} \\
\textit{ University of Birmingham}\\
Birmingham, United Kingdom \\
R.Stolkin@bham.ac.uk
}
}

\maketitle

\begin{abstract}
In this article, we study the problem of selecting a grasp pose on the surface of an object to be manipulated by considering three post-grasp objectives. These objectives include (i) kinematic manipulation capability \cite{ghalamzan2016task, ghalamzan2017human}, (ii) torque effort \cite{mavrakis2016analysis} and (iii) impact force in case of collision \cite{mavrakis2017safe} during post-grasp manipulative actions. 
In these works~\cite{mavrakis2016analysis,ghalamzan2016task, ghalamzan2017human,mavrakis2017safe}, the main assumption is that a manipulation task, i.e. trajectory of the centre of mass (CoM) of an object is given. In addition, inertial properties of the object to be manipulated is known. 
For example, a robot needs to pick an object located at point \emph{A} and place it at point \emph{B} by moving it along a given path. 
Therefore, the problem to be solved is to find an initial grasp pose that yields the maximum kinematic manipulation capability, minimum joint effort and effective mass along a given post-grasp trajectories. 
However, these objectives may conflict in some cases making it impossible to obtain the best values for all of them. 
We perform a series of experiments to show how different objectives change as the grasping pose on an object alters. 
The experimental results presented in this paper illustrate that these objectives are conflicting for some desired post-grasp trajectories. This indicates that a detailed multi-objective optimisation is needed for properly addressing this problem in a future work.
\end{abstract}

\section{Introduction}
\begin{figure}[tb!]
\centering
 \includegraphics[trim=0cm 0cm 0cm 0cm, clip=true,scale=.14,angle = 0]{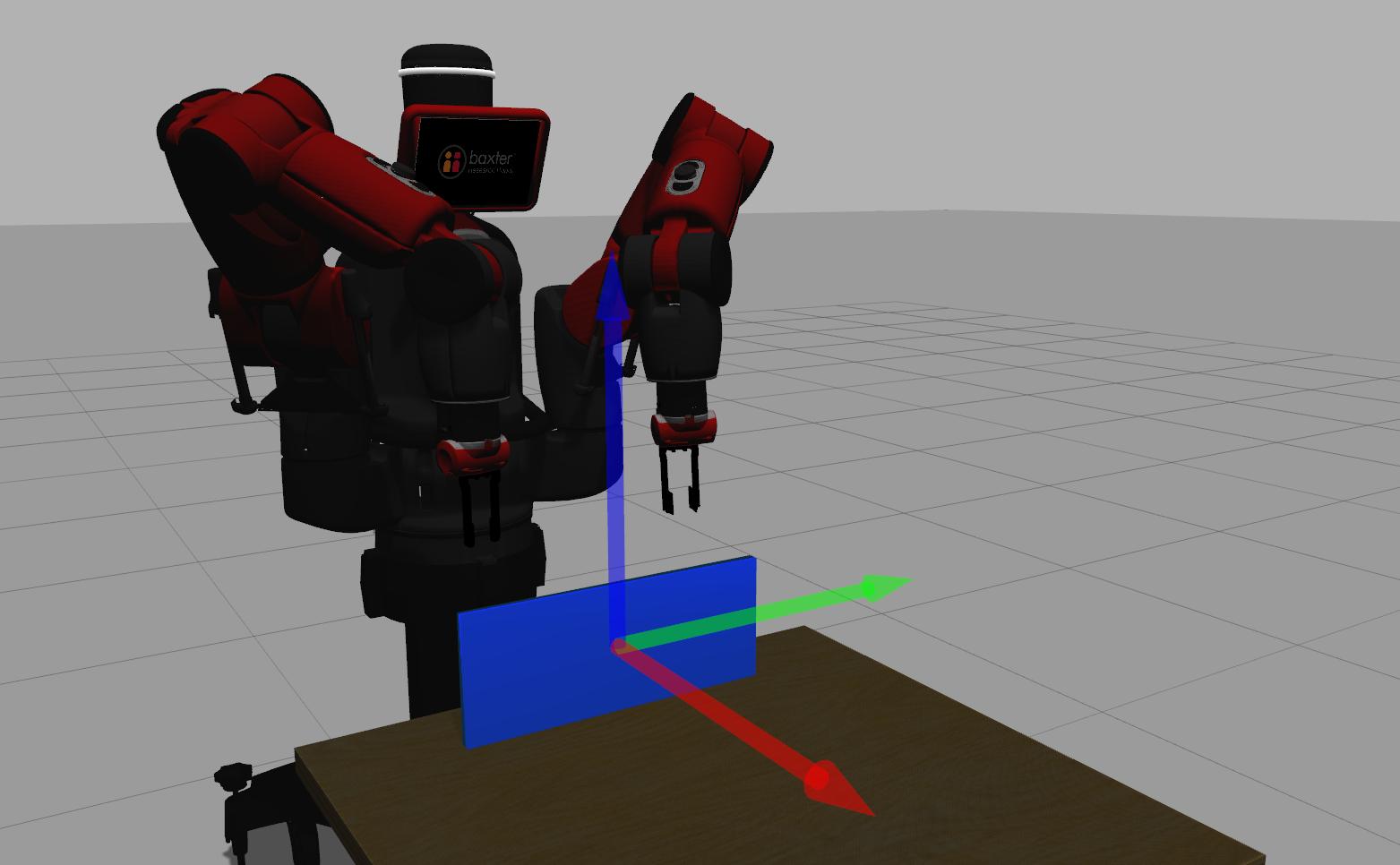}
\caption{Simulation setup with a Baxter robot in Gazebo simulator. The robot is tasked with grasping and then moving the object. This paper is concerned with enabling the robot to choose from several feasible grasps, to obtain some objectives during post-grasp actions. The robot manipulates a cuboid object with dimensions $0.5 \times0.15\times0.2\;[m^3]$ and mass of $0.4\;[kg]$. The coordinate axes of the object's centroid are shown, where red, green and blue correspond to $x$,$y$ and $z$ axes, respectively. The inertia tensor of the object is known in advance. 10 different grasps are generated for evaluation. We provide three different Pick-and-Place tasks for the robot to execute, and for each task and grasp, we pre-calculate the effective mass, the joint effort and the manipulability along the task trajectory. We aim to investigate the performance of each grasping point according to the calculated metrics. 
\label{fig:RealExp}}
\end{figure}

Grasping an object and performing manipulative actions are key distinguishing skills of primates~\cite{jeannerod1995grasping} learnt in early stage of skill development. 
In spite of the research conducted on this topic in different fields, e.g.  neuroscience~\cite{jeannerod1995grasping}, neuropsychology~\cite{iacoboni2005grasping}, this complex behaviour is not yet fully understood. 
Robotic grasping and manipulation have been widely inspired by the studies from other fields~\cite{lopes2007affordance}. 
Although a robot is desired to make stable contacts on object surface by its hand/fingers to move the object to another pose, most robotic grasping literature focuses on just computing contact points that make a stable force-closure \cite{ferrari1992planning}, or form closure grasp \cite{ding2001computation}. 
Hence, an obtained grasp pose may not be sufficiently good for manipulative motions after making stable contacts. 
The reason for this may be (i) computing a set of contact points on an object for a robotic hand using a 2D image or partial/full point cloud is yet an open challenging research question~\cite{kopicki2015one,saxena2008robotic,miller2004graspit} and (ii) planning the post-grasp trajectory~\cite{osa2017guiding, ratliff2009chomp} is a complex problem. Hence, they are often tackled in isolation. Nevertheless, this two complex problems must be jointly solved \cite{Cheraghpour2011} because they have dependent solutions, For example, to pick and object shown in Fig.~\ref{fig:RealExp} and to place it at the desired pose, the robot may grasp the object such that the torque effort/energy during manipulative actions becomes minimum. 

The lesson we learnt from our previous works~\cite{ghalamzan2016task,mavrakis2016analysis,mavrakis2017safe, ghalamzan2017human} suggest that post-grasp objectives can be efficiently optimised during reaching an object to grasp it. 
The main assumption in these works is that the trajectory of the centre of mass (CoM) of the object to be manipulated is given. 
These objectives can be used to prune a set of possible grasp pose candidates. 
These objectives, however, are task relevant (application example), i.e. one objective may become more important for an application example. 
For example, Ghalamzan et al.~\cite{ghalamzan2016task} proposed a kinematic velocity manipulability (TOV) and showed this criterion can be maximised by a suitable selection of grasp. 
TOV is integral of velocity manipulability along the desired manipulation trajectory. 
The grasp pose selected according to TOV has been shown to yield minimum joint velocities and consequently implicitly avoiding singularities along a post-grasp trajectory of object's CoM. 
Moreover, this metric is used in~\cite{ghalamzan2017human} to assist human teleoperating a manipulator for grasping an object.   
On the other hand, Mavrakis et al.~\cite{mavrakis2016analysis,mavrakis2017safe} studied the effect of grasp pose selection on torque effort and impact force along a post-grasp trajectory.

The ideas on ``task-informed grasp selection'' \cite{ghalamzan2016task,mavrakis2016analysis, mavrakis2017safe, ghalamzan2017human} poses this question that `` may the objectives, which are proposed for task-informed grasp selection, conflict?'' 
The answer to this question depends on the desired manipulative task and application example. For instance, in surgical robotics \cite{wilkening2017development}, minimising torque effort (energy consumption of the robot) may not be as important as minimising cognitive workload of the surgeon operator; hence, maximising task oriented kinematic manipulability becomes the main objectives for task-informed grasp selection \cite{ghalamzan2017human}. 
However, the objectives of task-informed grasp selection may conflict in many applications where they are all desired to be optimal.

In this paper, we study how the values of the objectives of task oriented grasp selection changes with a different selection of grasp poses. 
Our experimental results suggest a multi-objective optimisation is needed for optimally selecting a grasp pose in a future work.    

The remainder of this paper is as follows. First, in Section~\ref{Sec:OSTraj} problem formulation including our main assumption is presented. 
Next, the objectives of task-informed grasp selection are presented in Section~\ref{Sec:PGO}. Finally, experimental results in Section~\ref{Sec:exp} demonstrate that these objectives conflict showing that multi-objective optimisation is needed for better understanding optimal grasp selection.

\section{Operational space trajectory of a task}
\label{Sec:OSTraj}
\begin{figure}[!tp]
\centering
\begin{adjustbox}{width=.9\linewidth}
 \begin{tikzpicture}[x={(0.866cm,-0.5cm)}, y={(0.866cm,0.5cm)}, z={(0cm,1cm)}, scale=.6,
    >=stealth, %
    inner sep=0pt, outer sep=3pt,%
    wave/.style={very thick, color=#1,smooth},
    polaroid/.style={fill=black!40!white, opacity=0.3},
cube/.style={very thick,black!70},
			grid/.style={ thin,black!50},
			axis/.style={->,black,thick},
			axis1/.style={->,red,thick}]

	%draw a grid in the x-y plane
	\foreach \x in {-0.5,0,...,10}
		\foreach \y in {0,0.5,...,3}
		{
			\draw[grid] (\x,-0.0) -- (\x,3);
			\draw[grid] (-0.5,\y) -- (10,\y);
		}
	    \draw[blue,very thick,->] (1,1.25,0.5) -- (1,3 ,0.5) node[anchor=south ]{$x_1(t_1)$};
    \draw[blue,very thick,->] (1,1.25,0.5) -- (1.5,1.25,0.5) node[anchor= north ]{$z_1(t_1)$};
    \draw[blue,very thick,->] (1,1.25,0.5) -- (1,1.25,2+1) node[anchor=south east]{$y_1(t_1)$};

    % Frame at center of Mass({t_N})
    \coordinate (Oc) at (2, 1.25, 0.5);
    \draw[axis1] (Oc) -- +(2.5, 0,   0) node [below left] {$x_c(t_1)$};
    \draw[axis1] (Oc) -- +(0,  2.5, 0) node [right] {$y_c(t_1)$};
    \draw[axis1] (Oc) -- +(0,  0,   2.5) node [above right] {$z_c(t_1)$};       
    % Frame at Terminal point/center of Mass
    \coordinate (Oc2) at (7.75, 2.1, 1.25);
    \draw[axis1] (Oc2) -- +(2.5, 0,   0) node [below] {$x_c(t_N)$};
    \draw[axis1] (Oc2) -- +(0,  2.5, 0) node [right] {$y_c(t_N)$};
    \draw[axis1] (Oc2) -- +(0,  0,   2.5) node [above right] {$z_c(t_N)$};

	%draw the top and bottom of the cube
	\draw[cube] (1,1,0) -- (1,1.5,0) -- (3,1.5,0) -- (3,1,0) -- cycle;
	\draw[cube] (1,1,1) -- (1,1.5,1) -- (3,1.5,1) -- (3,1,1) -- cycle;
	
	%draw the edges of the cube
	\draw[cube] (1,1,0) -- (1,1,1);
	\draw[cube] (1,1.5,0) -- (1,1.5,1);
	\draw[cube] (3,1,0) -- (3,1,1);
	\draw[cube] (3,1.5,0) -- (3,1.5,1);
	    % monochromatic incident light with electric field
    \draw[wave=red!50, opacity=1, variable=\x, samples at={1,1.25,...,7.88}]
        plot (\x+1, {sin(1.0*(\x+.74) r)+0.25}, .03*\x^2-.03+0.5);        
		%draw the top and bottom of the cube
%	\draw[cube] (1,1,0) -- (1,1.5,0) -- (3,1.5,0) -- (3,1,0) -- cycle;
	\draw[cube] (7.7,.85,1.7) -- (7.7,1.35,1.7) -- (9.7,1.35,1.7) -- (9.7,.85,1.7) -- cycle;
	\draw[cube] (7.7,.85,2.7) -- (7.7,1.35,2.7) -- (9.7,1.35,2.7) -- (9.7,.85,2.7) -- cycle;
%	\draw[cube] (1,1,1) -- (1,1.5,1) -- (3,1.5,1) -- (3,1,1) -- cycle;
	
	%draw the edges of the cube
	\draw[cube]  (7.7,.85,1.7) -- (7.7,.85,2.7) ;
	\draw[cube] (7.7,1.35,1.7) -- (7.7,1.35,2.7);
	\draw[cube] (9.7,1.35,1.7) --  (9.7,1.35,2.7);
	\draw[cube]  (9.7,.85,1.7) -- (9.7,.85,2.7);			
    % Colors
    \colorlet{darkgreen}{green!50!black}
    \colorlet{lightgreen}{green!80!black}
    \colorlet{darkred}{red!50!black}
    \colorlet{lightred}{red!80!black}

    % Frame
    \coordinate (O) at (-.5, 0, 0);
    \draw[axis] (O) - - +(10.5, 0,   0) node [right] {${\pazocal{F}_g}$};
    \draw[axis] (O) - - +(0,  3.5, 0) node [above left] {${y_r}$};
    \draw[axis] (O) - - +(0,  0,   2.5) node [above] {${z_r}$};

            \draw[blue,very thick,->] (7.7,.85+.25,1.7+.5)    -- (7.7,2.75+.25,1.7+.5) node[anchor=south ]{$x_1({t_N})$};
    \draw[blue,very thick,->] (7.7,.85+.25,1.7+.5)    -- (8.2,.85+.25,1.7+.5) node[anchor=north ]{$z_1({t_N})$};
    \draw[blue,very thick,->] (7.7,.85+.25,1.7+.5)    -- (7.7,.85+.25,2.2+2.5) node[anchor=south east]{$y_1({t_N})$};
\end{tikzpicture}
\end{adjustbox}
\caption{An object in the global coordinate frame $\pazocal{F}_r =\{O_r,x_r,y_r,z_r\}$, shown in black. A local coordinate frame $\pazocal{F}_o =\{O_c,{x_c}{y_c},{z_c}\}$ is attached to the center of mass of the object, shown in red color. 
This frame follows a trajectory ${}^{c}\boldsymbol{\zeta}$ during manipulation. $\pazocal{F}_o ({t_1}) = \{O_c({t_1}),  x_c(t_1), y_c(t_1), z_c(t_1) \} $ and $\pazocal{F}_o ({t_N}) = \{O_c({t_N}),  x_c(t_N), y_c(t_N), z_c(t_N) \} $ denote this frame at the initial and terminal point of the manipulation trajectory with the corresponding frame of grasp candidate $\pazocal{F}_g ({t_1}) = \{O_g({t_1}),  {x_g({t_1})}, {y_g({t_1})}, {z_g({t_1})} \} $ and $\pazocal{F}_g ({t_N}) = \{O_g({t_N}),  {x_g({t_N})}, {y_g({t_N})}, {z_g({t_N})} \}$ shown with blue colour.
\label{Fig:Manipulation}}
\end{figure}
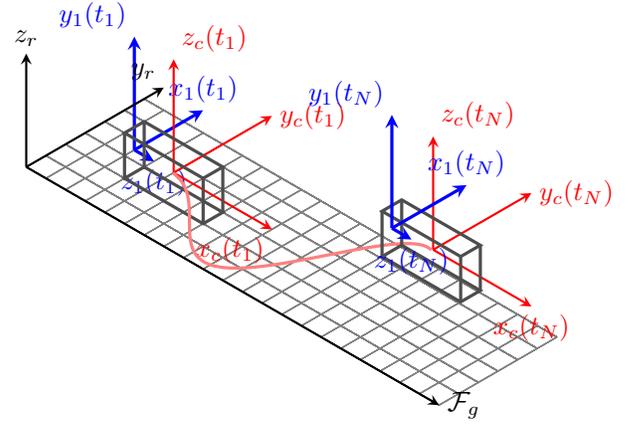

By operational point we refer to point $\pazocal{F}_g \in SE(3)$, attached to the end-effector, which will come into contact with the grasped object, once a successful grasp is achieved. 
$SE(3)$ denotes the group of 3D poses (3D position and 3D orientation).
Operational space trajectory $\boldsymbol{\zeta}_g$ refers to a vector of successive poses, of a frame attached to this point, defining a desired trajectory for the object.  
Let us denote a world reference frame $\{O_{r},{x_{r}},{y_{r}},{z_{r}}\}$ by $\pazocal{F}_g$.  
A trajectory to be followed by the manipulated object implies that local frame $\pazocal{F}_o$, attached to the CoM of the object, follows a sequence of poses: 
\begin{equation}
\begin{aligned}
\boldsymbol{\zeta}_c = \pazocal{F}_o(t)\\
0 \leq	t \leq \textrm{T}
\end{aligned}
\label{eq:ContTrej}
\end{equation}
where $t$ denotes a particular time during the motion, and $\textrm{T}$ is the total time that the robot needs to complete the desired manipulation task\footnote{Throughout this paper, $Y(t)$ denotes a continuous function of time, where $Y_i$ is a corresponding value of $Y(t)$ at time $t_i\;\forall\;i=1, . . ., n$, where $t_1 =0,\;t_n = \mathrm{T}$ and $0\leq t_i \leq \mathrm{T}$ denotes discrete sampling time. We also use $Y_t$ as a shorthand of $Y(t)$ where necessary.
${{}^*}\boldsymbol{\zeta}(t)$ and ${{}^*}{\zeta}(t)$ are continuous and discrete trajectory of poses of a frame attached to point $*$ of object in Figs.~\ref{Fig:Manipulation}.}.
$\pazocal{F}_o(t)$ determines a complete pose of the grasped object at time $t$. 
Although there are a variety of different possible representations of orientation, for the sake of simplicity here we use the conventional transformation matrix.

%<<<<<<<<<<<<<<<<<<<<<<<<   FIGURE   >>>>>>>>>>>>>>>>>>>>>>>>>>>
% \input{Figures/Manipulation}
%<<<<<<<<<<<<<<<<<<<<<<<<   FIGURE   >>>>>>>>>>>>>>>>>>>>>>>>>>>

Let us consider a local frame $\pazocal{F}_o=\{O_{c},{x_{c}},{y_{c}},{z_{c}}\}$. 
This frame can be described by a transformation matrix\footnote{ In general, ${}^{(..)}_{(.)}\mathbf{X}\in \mathbb{R}^{4\times4}$ denotes a transformation matrix from local frame $(.)$ into local frame $(..)$.} 
from the global reference frame $\{O_{r},{{}_{r}{x}},{y_{r}},{z_{r}}\}$ into the local frame $\{O_{c},{x_{c}},{y_{c}},{z_{c}}\}$:
%\footnote{${{}_c}X= \left[ \begin{array}{cccc}1 & 0&0\\0 & 1&0\\0 & 0&1\\1&1&1\end{array}\right]$ is a local frame fixed at the center of mass of the object, which represents zero rotation and translation w.r.t. itself.}: 
\begin{equation}
\begin{aligned}
{{}_r}T_o (t) =\left[
\begin{array}{c:c}
^r\mathbf{R}_o(t) &^r\mathbf{t}_o(t) \\ \hdashline
0_{1\times 3} & 1
\end{array}\right]_{4\times4}.
\end{aligned}
\label{eq:TransformationContinuous}
\end{equation}
% ==== Fig Grasp Candidate
Hence, $^r\mathbf{x}_o=\{^r\mathbf{t}_o,\,^r\mathbf{R}_o \}\in SE(3)$. Note that here we assume grasping and manipulation of rigid objects. 
%%%%%%%%%%%%%%%%%%%%%%%%%
% \input{Figures/LocalTransform}
%%%%%%%%%%%%%%%%%%%%%%%%%%
Let us denote a local frame attached to the robot end-effector, which we refer to as the ``operational point'', by $\pazocal{F}_g = \{ O_g,{x_g},{y_g},{z_g}\}$ which corresponds to the robotic arm configuration.
Because the object is non-deformable, any candidate robot wrist pose can be expressed by a fixed transformation matrix ${}_{o}T_g=\{^o\mathbf{t}_g, ^o\mathbf{R}_g\}$ from $\pazocal{F}_o$ into $\pazocal{F}_g$ %, which denotes a transformation from a frame of candidate pose ${{}^p}X$ in reference frame to local frame ${x_{c}_t}{y_{c}}_t{z_{c}_t}$
 (Fig.~\ref{Fig:Manipulation}):
 \begin{equation}
\begin{aligned}
^r\mathbf{R}_g(t) = {}^r\mathbf{R}_o(t){}^o\mathbf{R}_g\\
^r\mathbf{t}_g(t) = {}^r\mathbf{t}_o(t) +  {}^r\mathbf{R}_o(t) ^o\mathbf{t}_g
\end{aligned}.
\label{eq:GT}
\end{equation}
 For the sake of the simplicity of presentation, we choose to represent the orientation component of $\pazocal{F}_g$ with a quaternion parametrisation; hence, the trajectory of the end-effector is represented by $^r\mathbf{x}_g=\{^r\mathbf{t}_g, ^r\mathbf{\q}_g\}\in SE(3)$  where $\mathbf{\q}_g\in SO(3)$ represents the unit-quaternion associated to the rotation matrix $^r\mathbf{R}_g$. 
 %
%%%%%%%%%%%%%%%%%% Fig 1 Exp
\begin{figure}[tb!]
\centering
\subfigure[][]{    \includegraphics[trim=1cm 0.cm 2cm 0cm, clip=true,scale=.41,angle = 0]{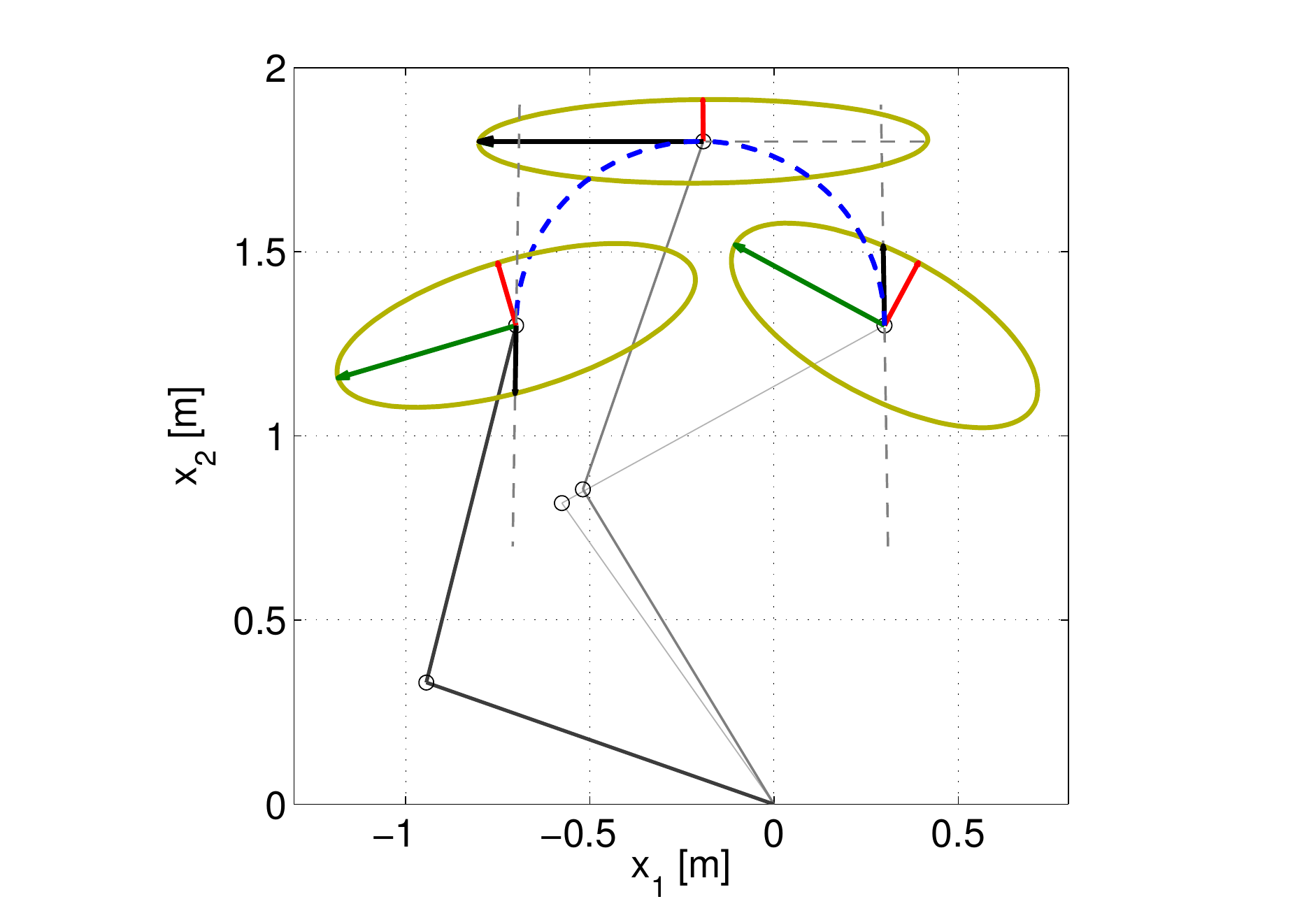}
  \label{fig:SimTOV}
}\hspace{\subfigtopskip}\hspace{\subfigbottomskip}
\subfigure[][]{
  \includegraphics[trim=0cm .cm 1cm 0cm, clip=true,scale=.41,angle = 0]{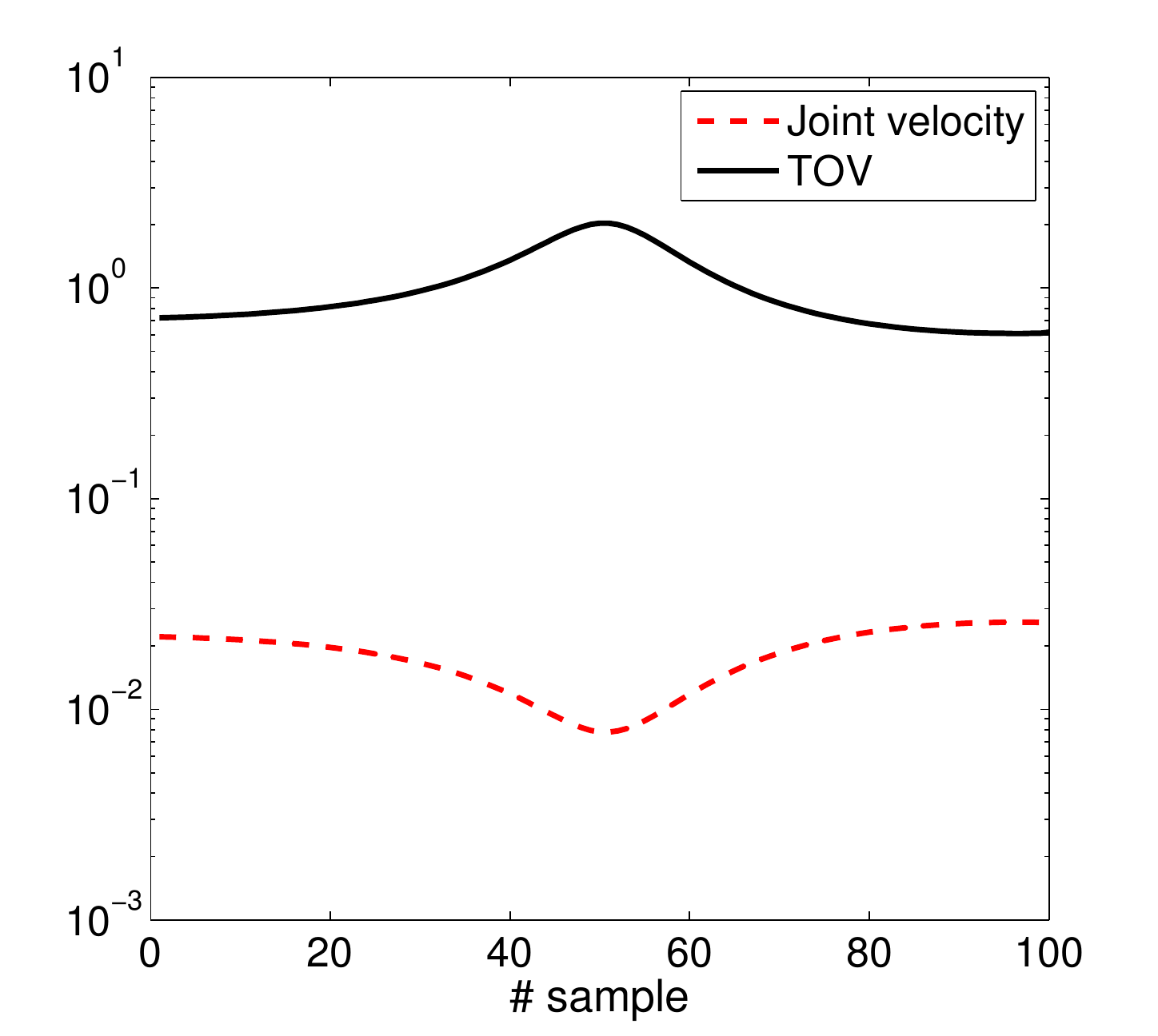}
  \label{fig:JV_TOV}
}
\caption{A 2-D manipulator follows a semi-circle from right to left shown with dashed blue line: \subref{fig:SimTOV} The manipulability ellipsoids are depicted at several configurations. Red and green arrows represent the ellipsoid major/minor axes. The proposed \emph{task oriented velocity manipulability} measure (black arrow) is obtained by evaluating the radius of the manipulability ellipsoid along the desired end-effector path. The corresponding TOV value and joint velocities are shown in \subref{fig:JV_TOV} against the samples' number of trajectory.
   \label{Fig:TM}}
   \vspace{-.2cm}
\end{figure}
%%%%%%%%%%%%%%%%%% Fig 1 Exp

\section{Post grasp objectives}
\label{Sec:PGO}
The problem of planning both grasps and subsequent manipulative actions have typically been studied in isolation. For example, $^o\mathbf{x}_g=\{^o\mathbf{t}_g, ^o\mathbf{\q}_g\} \subset \mathcal{X}_g$ is a set of wrist poses of the manipulator in eq.~\eqref{eq:GT}, which allow stable contacts of robots hand on object surface. Such poses can be computed by a variety of well known grasp planning algorithms (e.g. \cite{kopicki2015one,saxena2008robotic,miller2004graspit,ferrari1992planning,ding2001computation}, or other grasp-planners as the user prefers). The current state-of-the-art grasp planners, typically generate a set of various possible stable grasp configurations.  In \cite{kopicki2015one,saxena2008robotic} these are based on learned relationships, between features of the object's surface geometry and appropriate configurations of various parts of the robot hand. Alternatively, a set of potential stable grasps can be computed using grasp simulation software \cite{miller2004graspit}, force-closure analysis \cite{ferrari1992planning}, or form closure \cite{ding2001computation}. 
However, joint reasoning is essential to enable robots to both make a stable grasp and complete real manipulative tasks. 
In this paper, the two problems are considered jointly and a solution that takes both into consideration is proposed. Manipulative actions after making stable contacts may provide challenging research questions in different contexts. 
For instance, safety becomes very critical in the context of human-robot interaction. 
In this paper we consider three different criteria including (i) manipulation capability (in section \ref{TOKVM}), (ii) torque energy (in section \ref{TOKVM}) and (iii) impact force in the case of collision of robot hand with an object (in section \ref{TOKVM}) while performing post-grasp actions. We show that these objective values are functions of both a selected grasp pose and a post-grasp trajectory.

\subsection{Task Oriented Kinematic Velocity Manipulability (TOV)}  
\label{TOKVM}
Ghalamzan et al. \cite{ghalamzan2016task} introduced a task-relevant velocity manipulability cost function (TOV) to address the problem of jointly planning both grasps and subsequent manipulative actions. 
Later, this cost function is used in a mixed initiative, shared control for master-slave grasping and manipulation~\cite{ghalamzan2017human}. The novel system proposed in this work gives informative force cues to a human operator and assists her/him to grasp an object (making stable contacts between robot's hand and object surface) such that the TOV is maximum during manipulative actions. It is shown that maximising TOV results in significantly reduced joint velocities. 
\begin{comment}
\begin{figure}[tb!]%%% %%%%%
    \centering
    \includegraphics[trim=0.9cm 0.cm 0cm 0cm, clip=true,scale=.24,angle = 0]{with_force_without_force_grasp1}
  \caption{The object to be manipulated in the remote workspace and the given post-grasp trajectory (dotted red line). This figure shows the position from which the user grasped the object without haptic guidance (left) and with haptic guidance (right).}
   \label{Fig:two_grasp_poses}
\end{figure}
\end{comment}
Let $\btheta\in\mathbb{R}^6$ be the joint vector of the considered manipulator arm, and
\begin{equation}\label{eq:geom_Jacobian}
\u=\left[
\begin{array}{c}
\v_g \\ 
\bomega_g
\end{array}
\right]=\J(\btheta)\dot\btheta
\end{equation}
be the geometric Jacobian relating joint velocities to the end-effector linear/angular velocities $\u=(\v_g,\,\bomega_g)\in\mathbb{R}^6$ in the end-effector frame $\pazocal{F}_g$ (for ease of notation, we drop the superscript $g$ for the quantities in~(\ref{eq:geom_Jacobian})). 
Kinematic velocity manipulability ellipsoid is defined by 
\begin{equation}\label{eq:manip_ellipsoid0}
\u^T(\J\J^T)^{-1}\u=1
\end{equation}
that represents the capability of the robot manipulator in generating task space velocities for a given norm of joint velocities (thus, representing some sort of dexterity of the robot arm). In this work we are interested in maximising (in an integral sense) a particular \emph{task-oriented} manipulability measure derived from~(\ref{eq:manip_ellipsoid0}): the radius of the manipulability ellipsoid along the tangent vector to the desired path in task space. This is meant to ease as much as possible the execution of the desired trajectory~(\ref{eq:GT}) by the manipulator arm with the smallest possible control effort (norm of the joint velocities).

We consider $\btheta(t)$ being the trajectory in joint space associated to the end-effector trajectory and generated by the robot inverse kinematics~(\ref{eq:GT}) where $\u(t)$ is the corresponding linear/angular end-effector velocity at each time.  We decompose $\u(t)$ as $\u(t)=a(t)\bar\u(t)$, with $a(t)$ representing the norm of $\u(t)$ and $\bar\u(t)$ its (unit-norm) direction. From~(\ref{eq:manip_ellipsoid0}) it follows that, along the planned path,
\begin{equation}\label{eq:ellips_task}
a^2(t)\bar\u^T(t)(\J(\btheta(t))\J^T(\btheta(t)))^{-1}\bar\u(t)=1.
\end{equation}
It is easy to verify that the quantity $a(t)$ solution of~(\ref{eq:ellips_task}) represents the length of the ellipsoid radius along the direction $\bar\u(t)$, see also the illustrative example in Figs.~\ref{Fig:TM}. Our aim is to \emph{maximise} the quantity $a(t)$ along the whole path as defined in the following integral cost function:
\begin{equation}
\begin{aligned}
H_{\mathrm{TOV}}(^r\mathbf{x}_g)  = \int_{\boldsymbol{\zeta}_o} {a^2(^r\mathbf{x}_g,s)} \mathrm{d}s \\
= \int_{\boldsymbol{\zeta}_o}  \frac{1}{\bar\u^T(\J(\btheta)\J^T(\btheta)^{-1})\bar\u} \mathrm{d}s,
\label{eq:MAM_prel}
\end{aligned}
\end{equation}
where $0\leq s\leq 1$, $s$ is a parametrisation of the path, $s=0$ indicates $t=0$, $s=1$ shows $t=t_f$ and $t_f$ is the time to completion. From eq.~\eqref{eq:GT}, it can be confirmed that $\u= \u(^r\mathbf{x}_g,s)$, $\btheta= \btheta(^r\mathbf{x}_g,s)$. 
In \cite{ghalamzan2017human}, $H_{\mathrm{TOV}}$ is called Task-oriented velocity manipulability (TOV).

\subsection{Manipulator dynamics under load}
\label{TE}

In this section, we assume that the dynamic model of the robot is known to us, and we have the corresponding governing equation of motion of the manipulator in the joint space, as per Eq.~\eqref{eq:Dynamic}. Here, we are interested in computing the total energy consumption of the robot when executing the desired post-grasp trajectory. Hence, we need to obtain``augmented'' equation of motion, i.e. a combined equation of motion for both the robot and its grasped object, in the robot's joint space.

The joint space dynamic model of an $n$-degree-of-freedom (DOF) manipulator is defined by:
\begin{equation}\label{eq:Dynamic}
\textrm{M}(\btheta)\ddot{\btheta}+\textrm{C}(\dot{\btheta},\btheta)+\textrm{N}(\btheta) = \vec{\vec{\tau}}
\end{equation}
where $\btheta$ and  $\vec{\tau}\in \mathbb{R}^n$ are the vectors of joint positions and joint torques, respectively, and $\textrm{M}(\btheta)$ is the manipulator inertia matrix.

Again, it can be confirmed that $\btheta= \btheta(^r\mathbf{x}_g,s)$ and $\tau= \tau(^r\mathbf{x}_g,s)$ from eq.~\eqref{eq:GT}.
\begin{equation}\label{eq:Cori}
\textrm{C}_{ij}(\dot{\btheta},\btheta) = \frac{1}{2}\sum_{k=1}^n \left( \frac{\partial \textrm{M}_{ij}}{\partial \btheta_k} + \frac{\partial \textrm{M}_{ik}}{\partial \btheta_j} - \frac{\partial \textrm{M}_{kj}}{\partial \btheta_i}\right)\dot{\btheta}_k
\end{equation}
represent the Coriolis and centrifugal force terms.  
\begin{equation}\label{eq:Grav}
\textrm{N}(\btheta,\dot{\btheta}) = \frac{\partial V}{\partial \btheta}
\end{equation}
defines a gravitational force term, where $V(\btheta)$ is potential energy due to gravity. 
The dynamics of the robot in operational space are represented using the operational coordinate $\mathbf{x}$ as follows:
\begin{equation}\label{Dynamic}
{M}(\btheta)\ddot{\mathbf{x}}(t)+{C}(\dot{\btheta},\btheta) \dot{\mathbf{x}}(t)+{N}(\btheta) = F(t)
\end{equation}
where:
\begin{equation*}\label{Dyn_OS} 
{M} =  J^{-T}(\btheta) \textrm{ M}(\btheta) J^{-1}(\btheta),
\end{equation*}
$F = J^{-T}(\btheta)\vec{\tau}$, ${N}(\btheta,\dot{\btheta})$, ${C}(\dot{\btheta})$ are the corresponding gravitational and Coriolis terms in operational space, and $J(\btheta)$ is the robot's Jacobian.  Now, augmented dynamic model of manipulator and object to be manipulated can be computed using the generalised inertia matrix of an object  ${\textrm{M}_g}$.
\begin{equation*}\label{eq:InerTens}
{M}_o = 
\begin{pmatrix}
m{I}_{3x3} & {0} \\
{0} & {I}_{CoM}\\
\end{pmatrix} 
\end{equation*}
where $m$ and ${I}_{CoM}$ denote the object's mass and inertia tensor w.r.t. the CoM.
This inertia tensor can be expressed in $\pazocal{F}_g$ as follows:
\begin{equation}\label{eq:InerTenEffect}
{}^g{M}_{o}= E^{-T}(\mathbf{x}_g){M}_oE^{-1}(\mathbf{x}_g)
\end{equation}
where ${E(\mathbf{x}_g)}$ is the matrix transforming the linear and angular velocities of the object's CoM to generalised velocities in the frame attached to the end-effector. Accordingly, we represent the grasped object's dynamics in the joint space:
\begin{equation}\label{eq:InertTot}
\textrm{M}_{tot}(\btheta) = \textrm{M}_{arm}(\btheta) +\textrm{M}_{o}
\end{equation}
where $\textrm{M}_{o}=  \left[ J^T(\btheta) {M}_{o}(\mathbf{x}_g) J(\btheta)\right]$ is the grasped object's inertia tensor representation in the joint space. 

\subsubsection{Manipulation energy consumption}
We use $\textrm{M}_{tot}$ in eq.~\eqref{eq:Dynamic},~\eqref{eq:Cori} and~\eqref{eq:Grav} to compute the corresponding torque of augmented model of object and manipulator as per eq.~\eqref{eq:Dynamic}. 
Eventually, the energy consumption of the robot to manipulate the object along path $\mathbf{\zeta}_o$ is
\begin{equation}\label{EnCons}
H_{\mathrm{TME}} (^r\mathbf{x}_g) = \int_{\boldsymbol{\zeta}_o}\vec{\tau}^2\mathrm{d}s 
\end{equation}
\subsubsection{Effective mass definition}
While one can compute the force at every point of interest of the manipulator by writing the corresponding operational space equation, we can analyse the kinetic energy matrix $M(\btheta)$ and compute the impact force during a collision without needing to solve the second order differential equation in Eq.~\eqref{Dynamic}. 

It has been shown that a manipulator is perceived according to its effective mass during a collision (Eq.~\eqref{EffMass}), denoted by $m_e$.
In analogy, we define the effective mass of the total system as
\begin{equation}\label{EffMass}
m_e (^r\mathbf{x}_g,s)= \frac{1}{\bar{\u}^T M^{-1}_{tot}(x) \bar{\u}}
\end{equation}
where $M_{tot} = {}^gM_o+M$ expressed in the operational space and 
\begin{equation}\label{TEM}
H_{\mathrm{TEM}}(^r\mathbf{x}_g) = \int_{\boldsymbol{\zeta}_o}  m_e \mathrm{d}s 
\end{equation}
Ideally, a high value of $H_{\mathrm{TOV}}$ and small values of $H_{\mathrm{TEM}}$ and  $H_{\mathrm{TME}}$ are desired.  In a manipulation task, we would like to have minimum values of $H_{\mathrm{TEM}}$,  $H_{\mathrm{TME}}$ and $\frac{1}{ H_{\mathrm{TOV}}}$.
Although a native approach to minimise all can be achieved by an affine combination of all objectives, we will show that this approach is not sophisticated and the solution must be obtained through a multi objective optimisation approach.
% <<<<<<<<<<<<>>>>>>>>>>>>> 
 % <<<<<<<<<<<<>>>>>>>>>>>>> 
\begin{figure*}[tb!]
\centering
\subfigure[][]{    
\includegraphics[scale = 0.095]{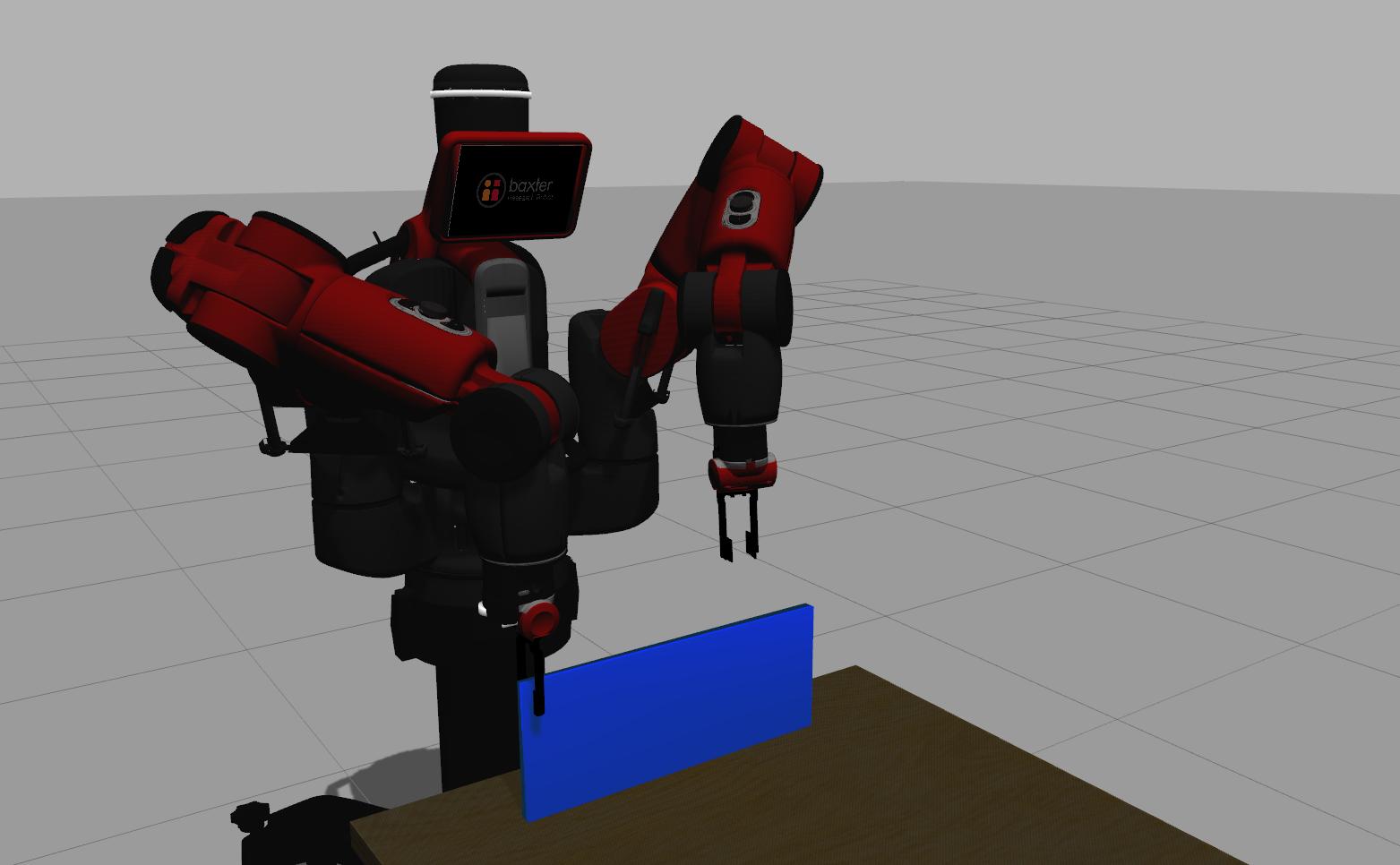}
  \label{fig:gp1}}
\hspace{\subfigtopskip}\hspace{\subfigbottomskip}
\subfigure[][]{
\includegraphics[scale = 0.095]{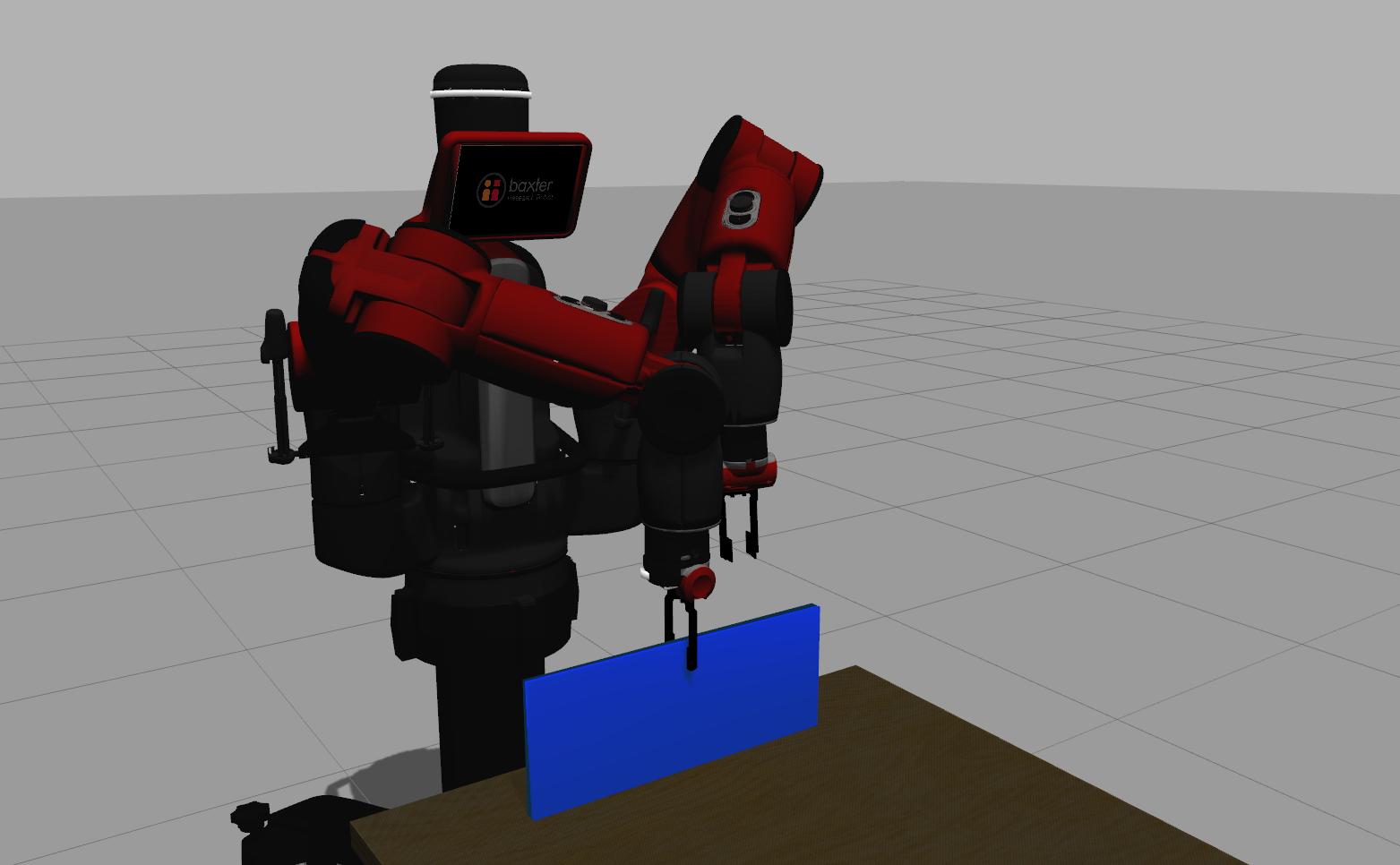}
\label{fig:gp2}}
\hspace{\subfigtopskip}\hspace{\subfigbottomskip}
\subfigure[][]{
\includegraphics[scale = 0.095]{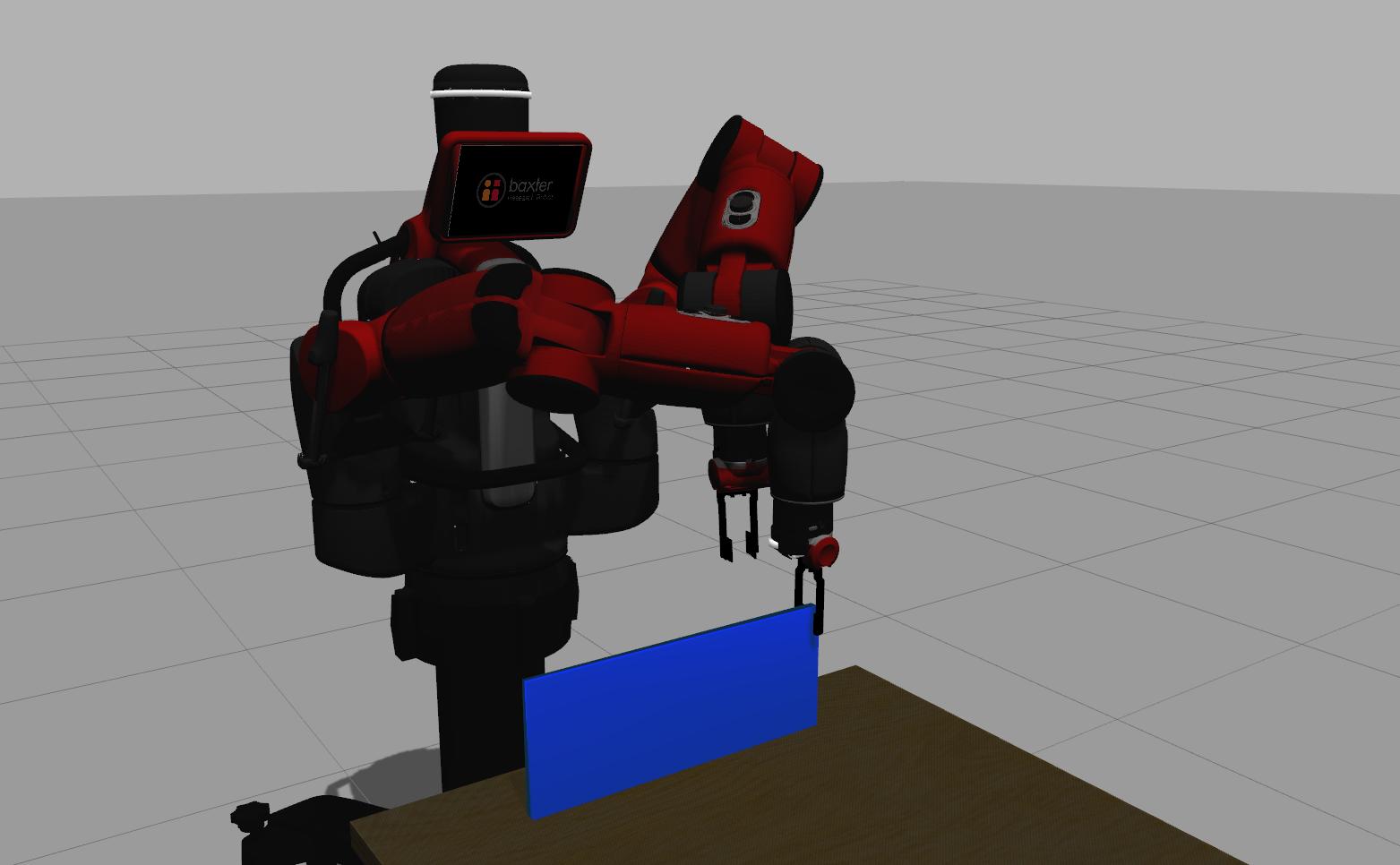}
\label{fig:gp3}}
\caption{In the first task, the Baxter is desired to pick up the object (blue cuboid), which is located on the table, move it $-20\: [cm]$ in line with y axis and $10\: [cm]$ in line with x axis and place it on the table. 
The x and y axes are shown with red and green arrows in Fig.~\ref{fig:RealExp}. 
All the 10 grasping poses are 
equally distributed on the top edge of the cuboid. Three example grasps on the object are shown where \subref{fig:gp1}, \subref{fig:gp2} and \subref{fig:gp3} show the first, fifth and tenth grasping pose.
\label{Fig::grasps}}
\end{figure*}
% <<<<<<<<<<<<>>>>>>>>>>>>> 

\section{Experimental results}
\label{Sec:exp}
To validate our hypothesis, i.e. performing a multi-objective optimisation is needed for selecting the best grasping pose, we conduct a series of experiments with a Baxter robot manipulating an object with a given task using the Gazebo simulator. 
The set-up is shown in Fig.~\ref{fig:RealExp}. The task is to pick a cuboid object and place it at different poses. The object has dimensions $0.5 \times 0.15\times 0.2\;[m^3]$ and uniform mass distribution with a mass value of $0.4\;[kg]$. We consider 10 different grasping poses on the object surface. The contact locations of the grasping poses are uniformly distributed on the top edge of the cuboid. Three of the generated grasp poses are shown in Fig.~\ref{Fig::grasps}. The first grasp is located at $-0.22\: [cm]$ and the last one is at $0.22\: [cm]$ along the y-axis. The Baxter approaches the contact points of each grasping pose on the top edge of the object from predefined approach points located $15\: [cm]$ above each grasping pose.

The robot is tasked with performing three Pick-and-Place: the robot lifts up the object $10\: [cm]$ from its initial position along the $z$ axis,
\begin{enumerate}
\item translates it in a combined motion $-20\: [cm]$ along the $y$ axis and $10\: [cm]$ along the $x$ axis and finally puts it down $-10\: [cm]$ along the $z$ axis;
\item translates $-0.35\: [cm]$ along the $y$ axis and finally puts it down on the table;
\item translates $-35\: [cm]$ along the $y$ axis and $-10\: [cm]$ along the $x$ axis and finally places it on the table.
\end{enumerate}
For the sake of comparison, we only present the results of pure translations.

\begin{figure}
\centering
\includegraphics[trim=0cm 0cm 0cm 2cm, clip=true, scale = 0.18]{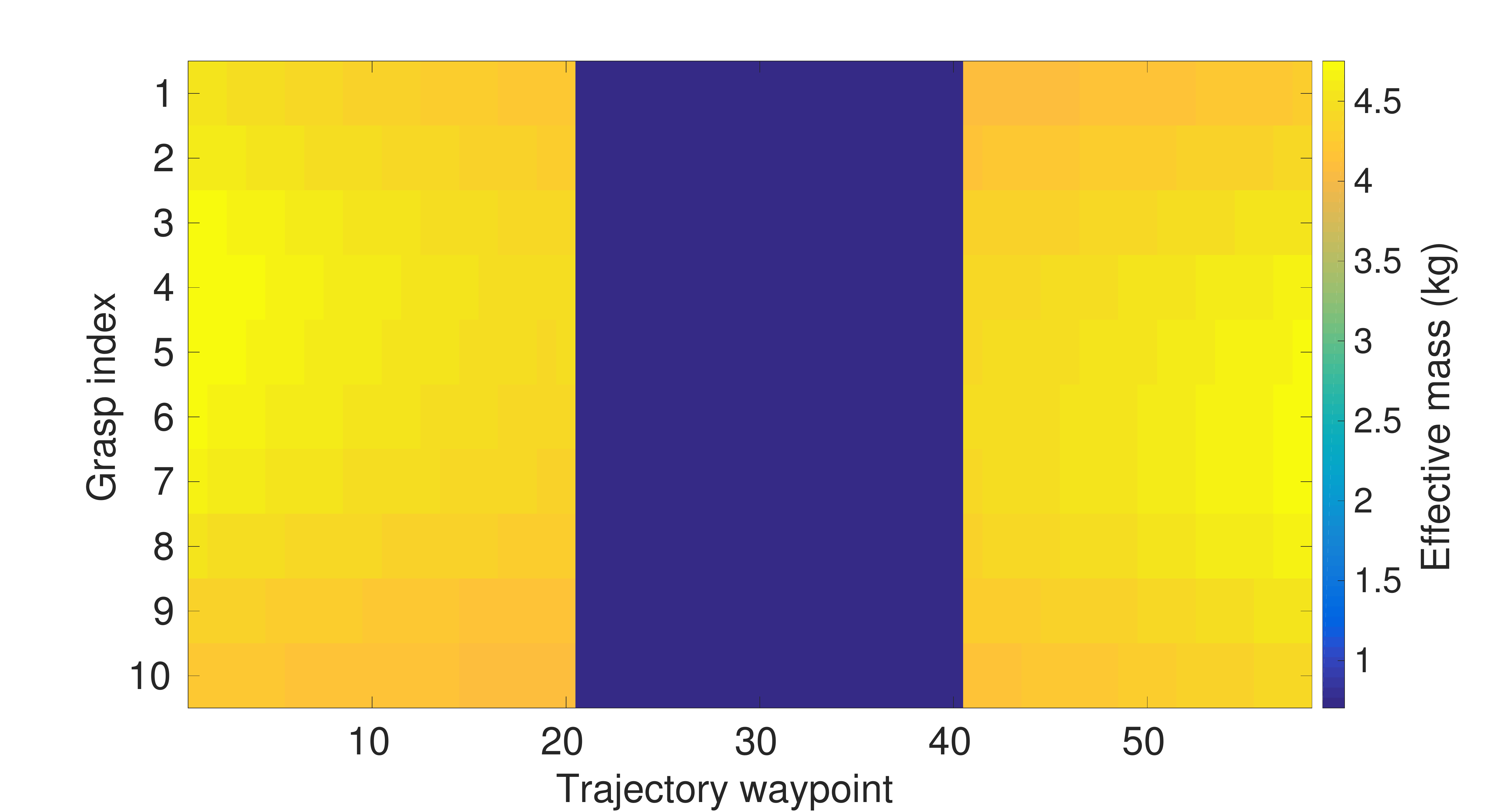}
\caption{Heat map of the computed effective mass for the third task. The horizontal axis represents the waypoints along the task trajectory and the vertical axis shows the grasp poses considered on the top edge of the object. This figure shows that the metric value of effective mass correlates with the waypoint of the pick-and-place trajectory and the selected grasp pose.   
\label{Fig::heatmap}}
\end{figure}

By using the Baxter PyKDL library, we compute the Jacobian and the dynamic model for each point of a trajectory allowing us to compute the metrics as per eq.~\eqref{eq:MAM_prel},~\eqref{EnCons},~\eqref{TEM}. 
An example of the effective mass for every initial grasping pose versus sample points of the third task trajectory is presented in Fig.~\ref{Fig::heatmap}. 
\begin{figure}[t!]
\centering
\subfigure[][]{    \includegraphics[trim=0cm .cm 0cm 0cm, clip=true, scale = 0.18]{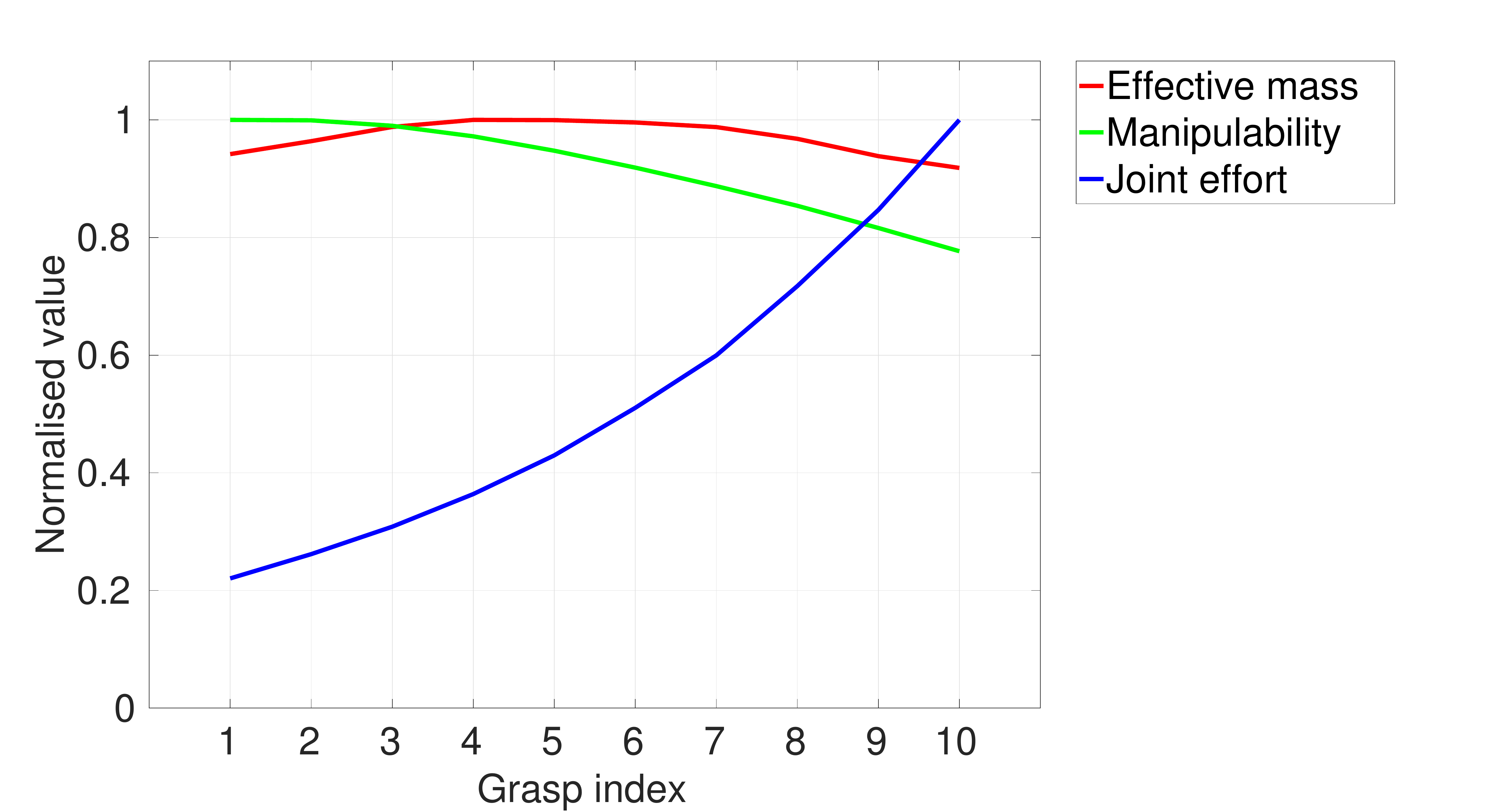}
  \label{fig:task1}}
\hspace{\subfigtopskip}\hspace{\subfigbottomskip}
\subfigure[][]{\includegraphics[trim=0cm .cm 0cm 0cm, clip=true,scale = 0.18]{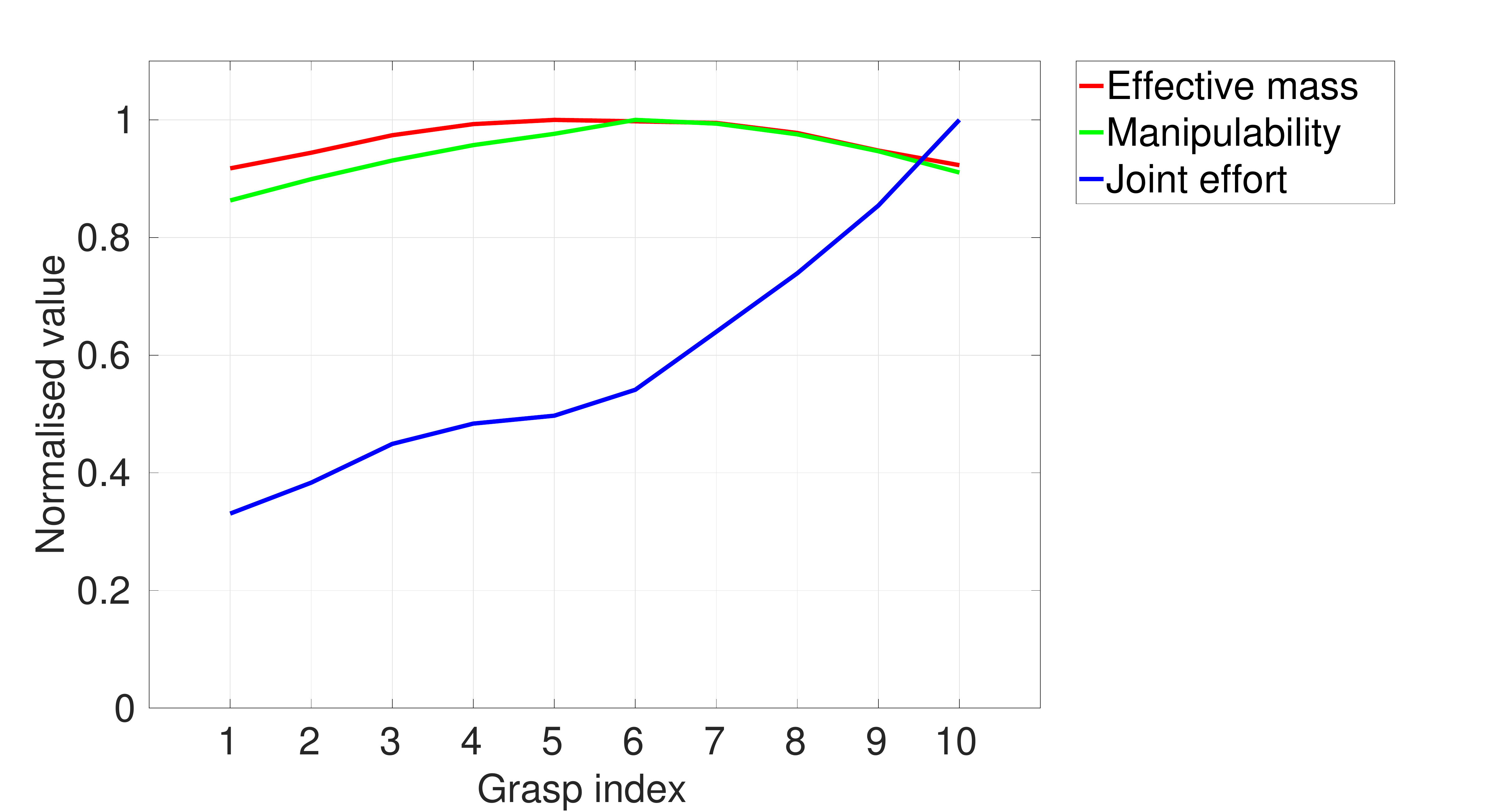}
\label{fig:task2}}
\hspace{\subfigtopskip}\hspace{\subfigbottomskip}
\subfigure[][]{\includegraphics[scale = 0.18]{task2}
\label{fig:task3}}
\caption{The final scalar metric values (namely TOV shown with green line, joint effort shown with blue line and effective mass shown with red line) for task 1 (top) 2 (middle) and 3 (bottom). The L2 norm of a metric along post-grasp trajectory yields a scalar value for each grasp pose. These values represent the quality of the grasp and are directly related to the task to be executed. As a result, the robot can choose a grasp that has low effective mass, low effort and higher manipulability. For instance, grasp number 1 in the first task, top figure,has maximum manipulability and minimum effective mass and effort.
\label{Fig::results}}
\vspace{-5pt}
\end{figure}

We computed the integrals presented in eq.~\eqref{eq:ellips_task}, \eqref{EnCons} and \eqref{TEM} for every tasks. 
For the sake of visualization, the metrics values of $\frac{1}{H_{\mathrm{TOV}}}$, $H_{\mathrm{TEM}}$ and  $H_{\mathrm{TME}}$ are normalised against their corresponding maximum values; that is,
\begin{equation*}
\begin{aligned}
\mathrm{H}_{\mathrm{TOV}} = \frac{ {H_{\mathrm{TOV}}} }{ \max\left({H_{\mathrm{TOV}}} \right)  },\\ 
\mathrm{H}_{\mathrm{TEM}} = \frac{H_{\mathrm{TEM}}}{\max\left(H_{\mathrm{TEM}} \right)}, 
\\
\mathrm{H}_{\mathrm{TME}} = \frac{H_{\mathrm{TME}}}{\max\left(H_{\mathrm{TME}} \right)}.
\end{aligned}
\end{equation*}
Theses normalised metrics are shown in Fig.~\ref{fig:task1},~\ref{fig:task2} and~\ref{fig:task3} for the first, second and third task.

A characteristic example is the first task (Fig.~\ref{fig:task1}), where we can see that the grasp No. 1 is the optimal yielding minimum effective mass, minimum joint effort and maximum manipulability. Furthermore, Fig.~\ref{fig:task1} shows the manipulability, effort and effective mass significantly changes with the choice of grasp poses. This enables us to use our methodology in choosing the grasp that is safe, yields the least effort and provides large manipulability for executing the task.

In contrast, the results yielded for the second task (Fig.~\ref{fig:task2}) shows that the objectives do not agree on the optimal grasping pose(Fig.~\ref{Fig::results}), i.e. while the effective mass and joint effort are implying that grasp number 1 is optimal, TOV manipulability suggests that grasp number 6 is optimal. 
Likewise, the indexes obtained for the task number 3 (Fig.~\ref{fig:task3}) shows they conflict, i.e. grasping pose number 2 yields minimum joint effort, whereas grasping pose number 1 is the best in terms of both TOV manipulability and effective mass. 

These results illustrate that the grasping pose selection for predefined manipulative actions is a complex multi-objective optimisation problem. Although one may consider an affine combination of these objectives for grasp selection, a more clever approach of multi-objective optimisation for grasp selection is needed which would be an interesting future work. 
% <<<<<<<<<<<<>>>>>>>>>>>>> 
% \input{exp.tex}

\section{Conclusion} 
\label{sec:conclusion}
Primates are capable of grasping and manipulating objects very efficiently by taking different objectives into consideration, e.g affordance of the object, maximum reachability and minimum wrist effort. 
In this paper, we presented an argument in favour of studying the problem of grasping pose selection as a multi-objective optimisation.  
In specific, we considered three cost functions of a given post-grasp trajectory presented in previous works \cite{ghalamzan2016task,ghalamzan2017human,mavrakis2016analysis,mavrakis2017safe} that have been used for selecting a grasp pose for a manipulator. These cost functions include (i) kinematic velocity manipulability (TOV) (ii) torque effort (the energy robot consumes to perform the manipulative actions) and (iii) impact force in the case the end effector of the manipulator collides with an obstacle. 
We presented a series of experiments. 
The results demonstrate how a manipulator can use the knowledge of post grasp actions and desired objectives for more intelligently grasping the object. 
Moreover, the results illustrate that the desired objectives (cost functions) conflict in some examples while they do not conflict each other in the first experiment. Our study suggests that a multi-objective optimisation must be used to better understand the problem of grasp selection according to the proposed objectives.

\end{document}